\documentclass[journal]{IEEEtai}
\usepackage[colorlinks,urlcolor=blue,linkcolor=blue,citecolor=blue]{hyperref}
\usepackage{graphicx,amsmath,booktabs,diagbox,array,multirow,multicol,subfigure,url,threeparttable,cite,bm,amsfonts,amssymb,threeparttable,hhline,bbding,makecell,bm,pifont,color}
\usepackage[table,xcdraw,dvipsnames]{xcolor}
\newcommand{\cmark}{\ding{51}}%
\newcommand{\xmark}{\ding{55}}%

\begin{document}

\title{Mixture-of-Experts for Open Set Domain Adaptation: A Dual-Space Detection Approach}

\author{Anonymous Authors}
\author{Zhenbang~Du, Jiayu~An, Yunlu~Tu, Jiahao~Hong, and~Dongrui~Wu,~\IEEEmembership{Fellow,~IEEE}
\thanks{Z. Du, J. An, Y. Tu, J. Hong and D. Wu are with the School of Artificial Intelligence and Automation, Huazhong University of Science and Technology, Wuhan 430074, China.}
\thanks{This work has been submitted to the IEEE for possible publication. Copyright may be transferred without notice, after which this version may no longer be accessible.}
\thanks{Dongrui Wu is the corresponding author. E-mail: drwu@hust.edu.cn.}}

\markboth{Journal of IEEE Transactions on ..., Vol. 00, No. 0, Month 2020}
{Mixture-of-Experts for Open Set Domain Adaptation: A Dual-Space Detection Approach}

\maketitle

\begin{abstract}
Open Set Domain Adaptation (OSDA) aims to cope with the distribution and label shifts between the source and target domains simultaneously, performing accurate classification for known classes while identifying unknown class samples in the target domain. Most existing OSDA approaches, depending on the final image feature space of deep models, require manually-tuned thresholds, and may easily misclassify unknown samples as known classes. Mixture-of-Experts (MoE) could be a remedy. Within a MoE, different experts handle distinct input features, producing unique expert routing patterns for various classes in a routing feature space. As a result, unknown class samples may display different expert routing patterns to known classes. In this paper, we propose \textbf{Dual-Space Detection}, which exploits the inconsistencies between the image feature space and the routing feature space to detect unknown class samples without any threshold. Graph Router is further introduced to better make use of the spatial information among image patches. Experiments on three different datasets validated the effectiveness and superiority of our approach. 
\end{abstract}

\begin{IEEEImpStatement}
  This paper proposes Dual-Space Detection (DSD) approach for Open Set Domain Adaptation (OSDA), which advances the field by accurately classifying known classes and identifying unknown classes without manually tuned thresholds. Leveraging the Mixture-of-Experts (MoE), DSD exploits inconsistencies between image and routing feature spaces to detect unknown samples. Additionally, our paper provides valuable insights on how to effectively make use of the routing feature in MoE. The introduction of the Graph Router further enhances the model's ability to utilize spatial information among image patches. Validated across three datasets, our method demonstrates superior performance over existing state-of-the-art OSDA techniques, offering a robust, scalable solution for real-world applications where data distribution and class labels vary.
  \end{IEEEImpStatement}


\IEEEpeerreviewmaketitle

\section{Introduction}

\IEEEPARstart{D}{eep} learning has made remarkable progress in image classification \cite{he2016deep,dosovitskiy2021an}. Nonetheless, most such models operate under the assumption that data are independently and identically distributed (i.i.d). In reality, factors such as variations in image styles and lighting conditions cause distribution shifts between source and target domain data \cite{zhao2020review, Deng2023TrBoosting}, frequently degrading the generalization.

\begin{figure}[htbp]    \centering
    \includegraphics[width=0.97\columnwidth]{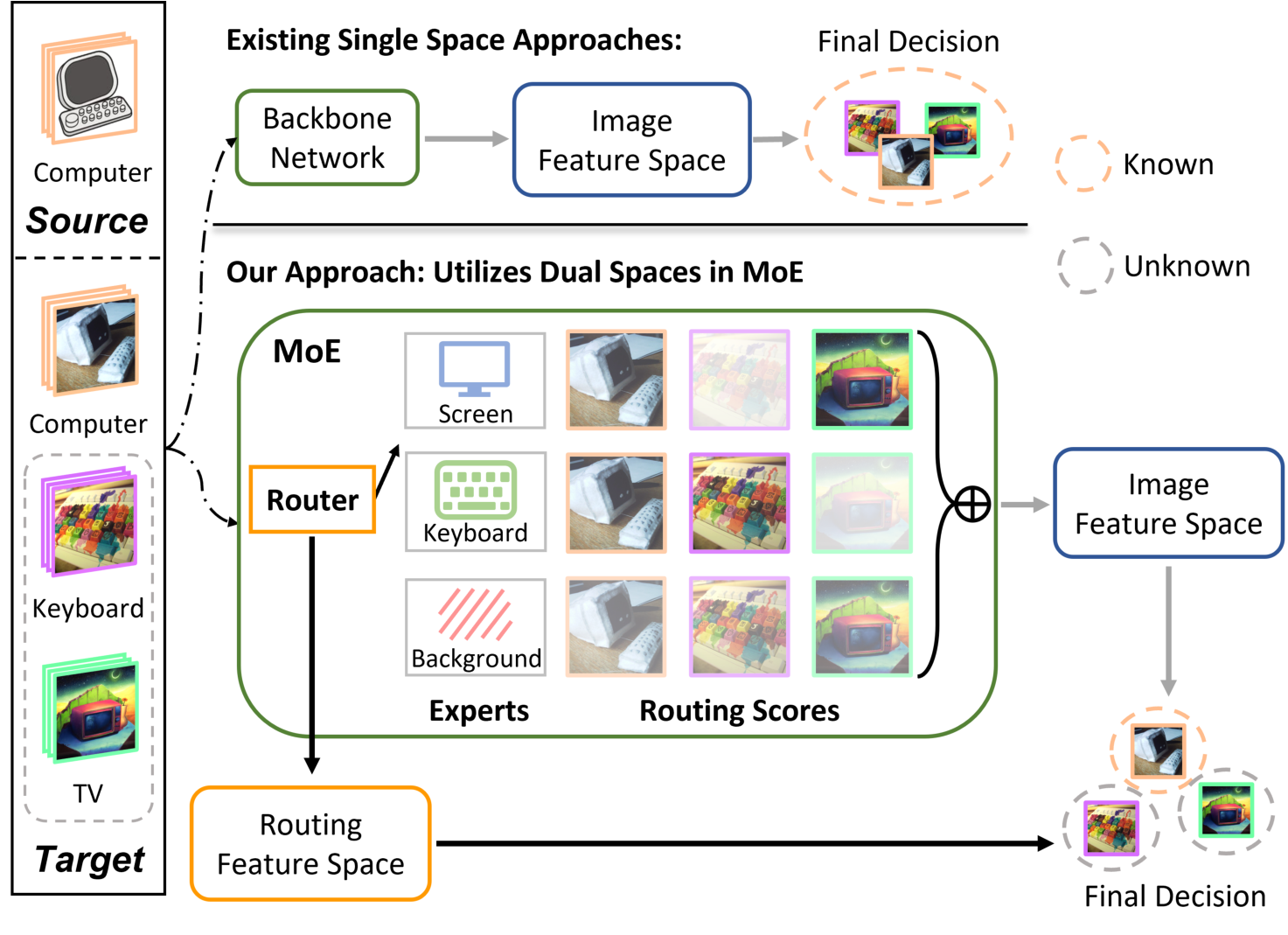}
    \caption{Comparison between existing OSDA approaches (top) and our proposed DSD (bottom). The transparency level of images corresponds to the degree of activation. DSD utilizes both the image feature space and the routing feature space in MoE to identify unknown samples, leading to improved performance.} 
    \label{Fig1}
\end{figure}

Unsupervised Domain Adaptation (UDA) \cite{ganin2015unsupervised, long2015learning} bridges this gap, enabling a model trained on the source domain to adapt to the target domain without using any target domain labels. Traditional UDA approaches, however, often assume that the source and target domains share identical classes. In many applications, the target domain contains unknown classes, i.e., classes unseen in the source domain. Open Set Domain Adaptation (OSDA) \cite{GHAFFARI2023101912, saito2018open, liu2019separate, bucci2020effectiveness, luo2020progressive, jang2022unknown, li2023adjustment} has been introduced to address this challenge. The goal is to handle the distribution and label shifts simultaneously, performing accurate classification for known classes while identifying unknown class samples in the target domain.

As depicted in Fig.~\ref{Fig1} (top), existing OSDA approaches predominantly use the image feature space derived from the backbone network, employing metrics like distances \cite{li2021domain}, classifier output probabilities \cite{saito2018open}, or entropy \cite{sanqing2023GLC}, to determine whether a target domain sample belongs to an unknown class or not. However, the unknown class samples might closely resemble known class samples, and the models trained on known classes may over-confidently misclassify an unknown class sample as known \cite{hendrycks2017a,dhamija2018reducing, li2023robustness}. Moreover, many existing approaches \cite{li2023adjustment, sanqing2023GLC} rely on carefully tuned thresholds, both as references and as training hyper-parameters, which may compromise the generalization and reliability of OSDA.

Mixture-of-Experts (MoE), consisting of multiple experts and a router, known for its ability to sparsify dense networks \cite{riquelme2021scaling, fedus2022review}, is an area of recent interest. Studies \cite{mustafa2022multimodal, li2023sparse} have shown that experts within the MoE demonstrate specific biases, handling distinct features within samples. The selection of these experts is governed by the router, which constructs a separate routing feature space determining the expert routing score for each sample, potentially providing essential insights to differentiate between classes.

This paper proposes an OSDA approach \textbf{Dual-Space Detection} (\textbf{DSD}), which extensively makes use of the routing feature space within the MoE, as illustrated in Fig.~\ref{Fig1} (bottom). We assume that known classes in the source and target domains should be similar to each other in both the image and routing feature spaces; however, unknown classes, even if they resemble known classes in the image feature space, may exhibit differences in the routing feature space. DSD identifies unknown class samples using information from both spaces, without any manually set thresholds. Concretely, target domain samples are assigned with pseudo-labels based on the distance to source domain class prototypes in both spaces, and samples with inconsistent pseudo-labels are clustered to obtain unknown class prototypes. Finally, all samples get closer to their nearest class prototypes via contrastive learning.

The router is a critical component of the MoE. Patch-wise routing \cite{riquelme2021scaling,fedus2022review, li2023sparse} is usually used in transformer-based MoE, which extracts routing features at the patch level instead of the sample-level (image-level), hindering the global critical information capture. For each image sample, different patches can be considered as nodes in a graph \cite{han2022vision}, and the adjacency relationships between different patches form the edges of the graph. Based on this, we propose a Graph Router that converts each image sample into a graph and uses a graph neural network \cite{kipf2017semisupervised, veli2018graph} to obtain a sample-level routing feature space.

Our main contributions include:
\begin{itemize}
    \item \textbf{Routing Feature:} We effectively utilize the routing feature in MoE, which has not been explored previously to the best of our knowledge.
    \item \textbf{Dual-Space Strategy:} We propose a novel threshold-free OSDA approach, DSD, to identify unknown class samples in the target domain, by exploiting the inconsistencies between the image feature space and the routing feature space in MoE.
    \item \textbf{Graph Router for MoE:} We introduce Graph Router, which uses a graph neural network as the MoE router to make better use of the spatial information in images.
    \item \textbf{State-of-the-Art Performance:} Our approach outperformed multiple baselines, including state-of-the-art universal domain adaptation and OSDA approaches, on three classical OSDA datasets.
\end{itemize}

The remaining of the article is organized as follows: Section \ref{sec:rela} introduces some related works. Section \ref{sec:meth} exhaustively describes our DSD and graph router.  Section \ref{sec:exp} shows our experimental results. Section \ref{sec:con} draws the conclusions and describes future research directions.

\section{Related Works}
\label{sec:rela}

This section reviews relevant literature in UDA, OSDA, and MoE.

\subsection{Unsupervised Domain Adaptation}
Generally, UDA techniques fall into three categories: discrepancy-based, self-supervised, and adversarial approaches. Discrepancy-based approaches \cite{long2015learning, sun2017correlation, lee2019sliced} minimize certain divergence metrics, which measure the difference between the source and target distributions. Self-supervised approaches \cite{ghifary2016deep, feng2019self, yue2021prototypical} typically incorporate an auxiliary self-supervised learning task to aid the model in capturing consistent features across domains and hence to facilitate adaptation. Adversarial approaches \cite{ganin2015unsupervised, long2018conditional, saito2019strong} leverage domain discriminators to promote domain-invariant feature learning. These approaches generally have difficulty in addressing label shifts between domains.

\subsection{Open Set Domain Adaptation}
OSDA handles discrepancies in known class distributions across the source and target domains, and simultaneously identifies target domain unknown classes. Various strategies have been proposed. OSDA by
Backpropagation (OSBP) \cite{saito2018open} trains a classifier for target instance classification and a feature extractor to either align known class samples or reject unknown ones. Rotation-based Open Set (ROS) \cite{bucci2020effectiveness} employs rotation-based self-supervised learning to compute a normality score for known/unknown target data differentiation. Progressive Graph Learning (PGL) \cite{luo2020progressive} integrates domain adversarial learning and progressive graph-based learning to refine the class-specific manifold. Unknown-Aware Domain Adversarial Learning (UADAL) \cite{jang2022unknown} uses a three-class domain adversarial training approach to estimate the likelihood of target samples belonging to known classes. Adjustment and Alignment (ANNA) \cite{li2023adjustment} considers the biased learning phenomenon in the source domain to achieve unbiased OSDA. Some Universal Domain Adaptation (UniDA) approaches, e.g., Domain Consensus Clustering (DCC) \cite{li2021domain}, leverages domain consensus knowledge to identify unknown samples, whereas Global and Local Clustering (GLC) \cite{sanqing2023GLC} adopts a one-vs-all clustering strategy. All these approaches use a single feature space, and often need some thresholds as hyper-parameters.

\subsection{Mixture-of-Experts}
A MoE model \cite{jacobs1991adaptive} includes multiple experts and a router. It aggregates the expert outputs via a weighted average, where the weights are determined by the router \cite{fedus2022review}. MoE models are sparse, hence reducing the computational cost and increasing the model capacity \cite{shazeer2017outrageously}, has been widely used in various fields \cite{shen2023mixture, du2023mixture, wu2023graphmetro, you2024shiftaddvit}.  The idea has also been extended to Vision Transformer (ViT) \cite{dosovitskiy2021an} to induce block sparsity \cite{riquelme2021scaling}. Studies \cite{mustafa2022multimodal, li2023sparse} have demonstrated MoE's promising performance in processing visual attributes. This paper improves the router of MoE in ViT and extends it to OSDA, using both the image feature space and the routing feature space to more effectively identify unknown class samples.

\section{Methodology}
\label{sec:meth}

\begin{figure*}[htpb]     \centering
    \includegraphics[width=\linewidth]{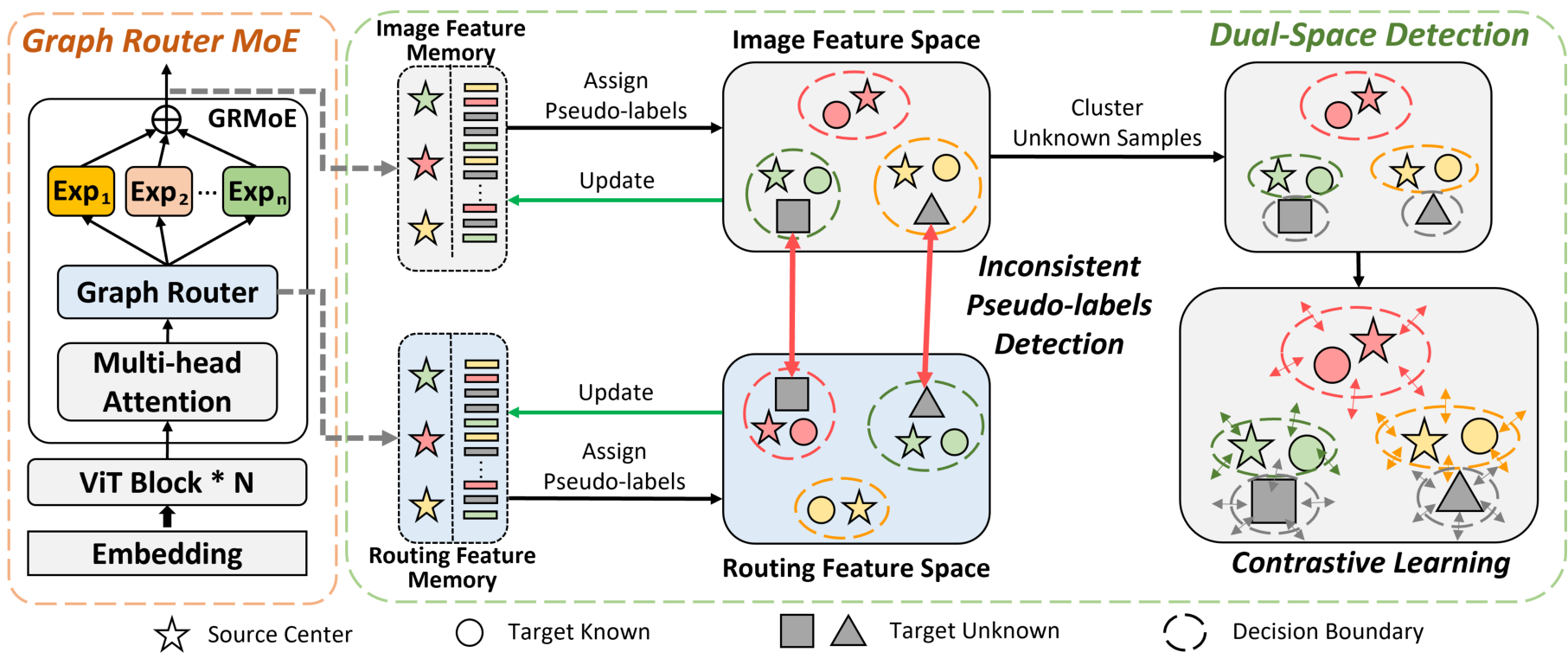}
    \caption{Graph Router MoE (left) and DSD (right). We store the model final output image features in the image feature memory, and the routing features (note they are different from the routing scores) in the routing feature memory. We then assign pseudo-labels to target domain samples in both spaces, and those with inconsistent pseudo-labels are clustered to obtain unknown class centers. Finally, we conduct contrastive learning on all samples and update both memory banks.}     \label{Fig2}     
    \end{figure*}
    
This section first introduces the formal definition of OSDA, and then our proposed DSD. Our code will be released after the acceptance. 

\subsection{Open Set Domain Adaptation}

OSDA deals with a labeled source domain and an unlabeled target domain. Assume there are $N_s$ labeled source domain samples $\mathcal{D}_s = \{(\bm{x}_i^s, y_i^s)\}_{i=1}^{N_s}$, where $\bm{x}_i^s$ is an input sample and $y_i^s$ its corresponding known class label, and $N_t$ unlabeled target domain samples $\mathcal{D}_t=\{\bm{x}_j^t\}_{j=1}^{N_t}$. Let $C_s$ be the label set in the source domain, and $C_s \cup y_{unk}$ the label set in the target domain, where $y_{unk}$ contains all unknown classes. The goal of OSDA is to accurately classify target domain samples from the known classes into the corresponding correct classes, and those from unknown classes into~$y_{unk}$.

For a given MoE model $G$ and an input $\bm{x}$, we use $\bm{f}$ to denote the image features from the last layer of $G$, and $\bm{r}$ the routing features (note they are different from the routing scores, as will be introduced in Section~\ref{sec:graph_router}).

\subsection{Overview}

Fig.~\ref{Fig2} illustrates our proposed DSD and Graph Router.

Fig.~\ref{Fig2} (left) shows a modified ViT model \cite{dosovitskiy2021an}, where a Graph Router MoE (GRMoE) layer is proposed to replace some conventional layers in the traditional ViT. More details will be presented in Section~\ref{sec:graph_router}.

Fig.~\ref{Fig2} (right) shows the flowchart of DSD. First, we feed the source and target domain samples into the source domain pre-trained model to obtain representations that will be stored in the image feature memory and the routing feature memory for source class prototypes (Section~\ref{sec:known}) and target domain samples (Section~\ref{sec:memory}), respectively. Next, we assign pseudo-labels to all target samples, based on their similarities to the source class prototypes in both the image feature space and the routing feature space. We then cluster target samples with inconsistent pseudo-labels between these two spaces to obtain class prototypes for potential unknown classes (Section~\ref{sec:prototype}). Finally, the model is updated by minimizing a contrastive loss between all  samples and all class prototypes, and the memory banks are updated accordingly (Section~\ref{sec:learning}).

\subsection{Graph Router MoE} \label{sec:graph_router}

GRMoE is shown in Fig.~\ref{Fig2} (left). One ViT encoder block consists of a Multi-Head Self-Attention (MHSA) and a Feedforward Network (FFN) with shortcut connections \cite{dosovitskiy2021an}. Sparsifying the FFNs in ViT improves domain generalization, as it maps features into a unified space so that similar features from different domains are processed by the same experts \cite{li2023sparse}. The router maps samples into a routing feature space, which provides important information for identifying the unknown samples. We use a graph neural network \cite{kipf2017semisupervised, veli2018graph, Fey2019wv} as the router to enhance the model's ability to utilizing spatial information.

\begin{figure}[htpb]    \centering
  \includegraphics[width=1.0\columnwidth]{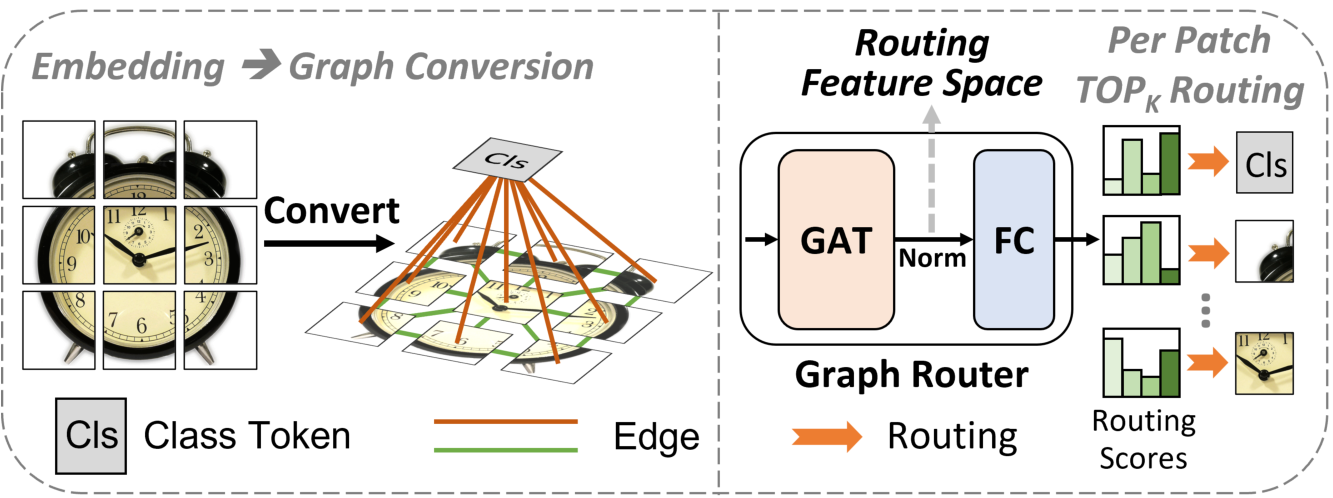}
  \caption{Overview of the Graph Router. (Left) The conversion from the embeddings to the graph. Each patch embedding serves as a node. The edges are formed by connecting adjacent patches of the original image and linking every patch to the class token. (Right) The graph is input into the Graph Router. The routing features are extracted from the GAT layer, and the routing scores are obtained from the FC layer. `Norm' denotes the normalization operation.}     \label{Fig3} 
\end{figure}

In a GRMoE layer, the FFN is replaced by an MoE, and each expert $\text{Exp}(\cdot)$ is implemented by an FFN \cite{riquelme2021scaling, li2023sparse}. The output of GRMoE is:
\begin{align}\label{eq:moe1}
    \text{GRMoE}(\bm{x}) = \sum_{i=1}^N \text{TOP}_K \{ \text{Softmax}(\text{GR}(\bm{x})) \} \cdot \text{EXP}_i (\bm{x}),
\end{align}
where $N$ is the total number of experts, and $\bm{x}$ the output of the MHSA, including a series of image patch embeddings and a class token. $\text{TOP}_K(\cdot)$ is a one-hot encoding that sets all other elements in the output vector as zero except the largest $K$ elements. $\text{GR}(\cdot)$ is our proposed Graph Router:
\begin{align}\label{eq:moe}
    \text{GR}(\bm{x})=\text{FC} ( \text{Norm} (\text{GAT} (\mathcal{G}(\bm{x})))),
\end{align}
where $\bm{x}$ is mapped into a graph $\mathcal{G}(\bm{x}) =(\mathcal{V}_{\bm{x}}, \mathcal{E}_{\bm{x}})$. The node set $\mathcal{V}_{\bm{x}}$ is the collection of the embeddings and the class token \cite{han2022vision}. The edge set $\mathcal{E}_{\bm{x}}$ is formed by connecting adjacent patches of the original image and linking every patch to the class token, as illustrated in Fig.~\ref{Fig3} (left). The graph is then input into the Graph Attention Layer \cite{veli2018graph} $\text{GAT}(\cdot)$, which is normalized by $\text{Norm} (\cdot)$ to obtain per patch routing features. The class token is used as the routing feature of the sample as it incorporates information from all other patches. The routing features are then input into a Fully Connected layer $\text{FC}(\cdot)$ to assign routing scores to the patchs and class token, as shown in Fig.~\ref{Fig3} (right). The output of the GRMoE layer is a summary of $\text{TOP}_K$ experts outputs weighted by their routing scores.

\subsection{Known Class Prototypes} \label{sec:known}

For source domain samples $\mathcal{D}_s$ from known classes $C_s$, since their labels are available, the class prototypes are obtained by feeding the source samples into the model and aggregating the final image features as well as the routing features from the router.

Specifically, the prototype of each known class is computed as the mean feature of all samples belonging to it:
\begin{align}
	\bm{c}_{k}^{r} = \frac{1}{|{\cal R}_{k}|}\sum_{\bm{r}_i \in {\cal R}_{k}} \bm{r}_i, \\
    \bm{c}_{k}^{f} = \frac{1}{|{\cal F}_{k}|}\sum_{\bm{f}_i \in {\cal F}_{k}} \bm{f}_i,
\end{align}
where ${\cal R}_{k}$ and ${\cal F}_{k}$ denote class $k$'s routing features and image features, respectively.

\subsection{Momentum Memory} \label{sec:memory}

We introduce a momentum memory \cite{ge2020selfpaced} for both the image feature space and the routing feature space to stabilize the learning. In each iteration, the encoded feature vectors in each mini-batch are used to update the momentum memory.

The prototype $\{\bm{c}^{r}_k, \bm{c}^{f}_k\}$ for the $k$-th source domain known class is updated as:
\begin{align} \label{eq:m_cen}
\bm{c}^{r}_k \leftarrow m \bm{c}^{r}_k + (1-m)\cdot \frac{1}{|\mathcal{B}_k^r|}\sum_{\bm{r}_i \in \mathcal{B}_k^r} \bm{r}_i, \\
\bm{c}^{f}_k \leftarrow m \bm{c}^{f}_k + (1-m)\cdot \frac{1}{|\mathcal{B}_k^f|}\sum_{\bm{f}_i \in \mathcal{B}_k^f} \bm{f}_i,
\end{align}
where $\mathcal{B}_k^r$ and $\mathcal{B}_k^f$ denote the set of source routing features and image features belonging to class $k$ in the current mini-batch, and $m \in [0,1]$ is a momentum coefficient.

The target domain samples do not have real labels, so we consider each sample as a separate class. Assume the $i$-th sample in the entire target domain is corresponding to the $i'$-th sample in the current mini-batch. Then, its features $\{\bm{r}^t_{i}, \bm{f}^t_{i}\}$ are updated as:
\begin{align}
\bm{r}_{i}^t &\leftarrow m\bm{r}_{i}^t + (1-m)\bm{r}^t_{i'}, \\
\bm{f}_{i}^t &\leftarrow m\bm{f}_{i}^t + (1-m)\bm{f}^t_{i'}. \label{eq:m_ins}
\end{align}

\subsection{Unknown Class Prototypes} \label{sec:prototype}

For an unlabeled target sample $x_j^t$, we first obtain its image features $\bm{f}^{t}_j$ and routing features $\bm{r}^{t}_j$ from the memory banks, and then compute the corresponding cosine distance $d(\cdot, \cdot)$ to each known class prototype. Two pseudo-labels $\hat{y}^{r}_j$ and $\hat{y}^{f}_j$ are next computed as:
\begin{align}
	\hat{y}^{r}_j =  \mathop{\arg\min}\limits_{k}( d(\bm{r}^{t}_j, \bm{c}_{k}^{r})),\\
  \hat{y}^{f}_j =  \mathop{\arg\min}\limits_{k}(d(\bm{f}^{t}_j, \bm{c}_{k}^{f})).
\end{align}

The pseudo-label $\hat{y}_j$ of $x_j^t$ is obtained by:
\begin{align}
    \hat{y}_j &= \left\{
    \begin{aligned}
     &\text{unknown}, &\text{if $\hat{y}^{r}_j \neq \hat{y}^{f}_j$},\\
     &\hat{y}^{f}_j, &\text{if $\hat{y}^{r}_j = \hat{y}^{f}_j$}.
    \end{aligned}
    \right.
    \end{align}

After obtaining pseudo-labels of all target samples, those with `unknown' pseudo-labels are then clustered into $n_u$ clusters using \textit{k}-means clustering. The Silhouette criterion \cite{rousseeuw1987silhouettes} is used to determine the optimal $n_u$. Given a sample $x$ in cluster $\mathcal{P}_I$, its Silhouette score $s(x)\in[-1,1]$ is computed as:
\begin{equation}
    \begin{aligned}
     a(x) &= \frac{1}{|\mathcal{P}_I| - 1}\sum_{x_i \in \mathcal{P}_I,  x_i \ne x} d(x, x_i),\\
     b(x) &= \min_{J\neq I}\frac{1}{|\mathcal{P}_J|}\sum_{x_i\in \mathcal{P}_J} d(x, x_i),\\
     s(x) &= \frac{b(x) - a(x)}{\max\{a(x), b(x)\}},
\end{aligned}
\end{equation}
where $a(x)$ is the average distance of $x$ to other members in the same cluster $\mathcal{P}_I$ (indicating cohesion), and $b(x)$ its nearest average distance to another cluster $\mathcal{P}_{J\neq I}$ (indicating separation). A high $s(x)$ means that $x$ aligns well within its own cluster and is distinct from all other clusters. We find the $n_u$ with the highest Silhouette score from $\{0.25|C_s|, 0.5|C_s|, 1.0|C_s|,2|C_s|,3|C_s|\}$, as in \cite{li2023sparse}.

After clustering, each cluster serves as a pseudo unknown class. The $k$-th unknown class prototype is computed as the mean image feature of all samples in its cluster:
\begin{align}
    \bm{w}_{k}^{f} = \frac{1}{|{\cal P}_{k}|}\sum_{\bm{f}^t_i \in {\cal P}_{k}} \bm{f}^t_i,\quad k=1,...,n_u,
\end{align}
where ${\cal P}_{k}$ denotes cluster $k$'s image feature set. 

\subsection{Training Loss} \label{sec:learning}

Considering the classification ability of the model and the balanced use of different experts in MoE, the training loss is:
\begin{align}
    \mathcal{L} = \mathcal{L}_{\text{con}} + \gamma \mathcal{L}_{\text {blc}}, \label{eq:gamma}
\end{align}
where $\mathcal{L}_{\text{con}}$ is the contrastive loss which drives each sample closer to the prototype of its most similar class and further away from the prototypes of the others, $\mathcal{L}_{\text {blc}}$ a regularization term to encourage balanced use of experts in MoE \cite{shazeer2017outrageously, li2023sparse}, and $\gamma$ a hyper-parameter.

Specifically, the contrastive loss is defined as:
\begin{align}
    \mathcal{L}_{\text{con}} = - \log \frac{\exp{(\langle \bm{f}, \bm{z^{+}} \rangle  )}}{\sum_{k=1}^{|C_s|} \exp{(\langle \bm{f}, \bm{c}^f_k \rangle  )} + \sum_{k=1}^{n_u} \exp{(\langle \bm{f}, \bm{w}_{k}^f \rangle )}},
    \label{eq:joint}
\end{align}
where $\bm{f}$ is an image feature vector, $\bm{z^{+}}$ the nearest class prototype to $\bm{f}$, and $\langle\cdot,\cdot\rangle$ the inner product between two image feature vectors to measure their similarity.

The balanced loss $\mathcal{L}_{\text{blc}}$ is used to mitigate the potential imbalance among the experts in MoE \cite{shazeer2017outrageously, mustafa2022multimodal, li2023sparse}:
\begin{align}
    \mathcal{L}_{\text {blc}} = \frac{1}{2} (\mathcal{L}_{\text{imp}}  + \mathcal{L}_{\text{load}} ).
\end{align}

The first term, $\mathcal{L}_{\text{imp}}$, encourages a uniform distribution of routing weights among all experts. The importance of the $i$-th expert is calculated as the sum of $\text{GR}_i(\bm{x})$ (the $i$-th component of the router output) for a batch $X$:
\begin{align}
\label{eq:importance}
    \text{imp}_i(X)= \sum\limits_{\bm{x} \in X} \text{GR}_i(\bm{x}).
\end{align}
$\mathcal{L}_{\text{imp}}$ is then computed by the squared coefficient of variation of $\{\text{imp}_i(X)\}^N_{i=1}$:
\begin{align}
   \mathcal{L}_{\text{imp}} = \left(\frac{\text{Std}(\{\text{imp}_i(X)\}^N_{i=1})}{\text{Mean}(\{\text{imp}_i(X)\}^N_{i=1})}\right)^{2}.
\end{align}
in which $\text{Std}(\cdot)$ is the standard deviation and $\text{Mean}(\cdot)$ is the mean.

The second term, $\mathcal{L}_{\text{load}}$, encourages balanced assignment across different experts. The probability that the $i$-th expert still remains in the selected $K$ experts after random noise re-sampling is:
\begin{align}
    p_{i}(x) = 1 - \Phi\left(\frac{\eta_K - \text{Softmax}_{i}(\text{GR}(\bm{x}))}{\sigma}\right), \label{eq:load_prob}
\end{align}
where $\eta_{K}$ is the smallest entry among the largest $K$ routing scores, $\sigma$ the sum of noise-disrupted routing scores, and $\Phi$ the cumulative distribution function of a Gaussian distribution. The load of the $i$-th expert for a batch $X$, $\text{load}_i(X)$, is defined as the sum of these probabilities for all samples in that batch:
\begin{align}
\label{eq:importance}
    \text{load}_i(X)= \sum\limits_{\bm{x} \in X} p_i(\bm{x}).
\end{align}

Then, $\mathcal{L}_{\text{load}}$ is computed by:
\begin{align}
   \mathcal{L}_{\text{load}} = \left(\frac{\text{Std} (\{\text{load}_i(X)\}^N_{i=1})}{\text{Mean} (\{\text{load}_i(X)\}^N_{i=1})}\right)^{2}.
\end{align}

\subsection{Inference}

With the known class prototypes $\{\bm{c}^f_1, \bm{c}^f_2,...,\bm{c}^f_{|C_s|}\}$ and unknown class prototypes $\{\bm{w}_{1}^{f},\bm{w}_{2}^{f},...,\bm{w}_{n_u}^{f}\}$ from training, to perform inference for each input target domain sample, we simply identify the nearest class prototype and use its label. Note that if any of $\{\bm{w}_{1}^{f},\bm{w}_{2}^{f},...,\bm{w}_{n_u}^{f}\}$ is chosen as the nearest prototype, then a label $y_{unk}$ is assigned to the corresponding input target sample.

\section{Experiments}
\label{sec:exp}

This section evaluates the performance of DSD in OSDA.

\begin{table*}[htpb]   \centering
\caption{Results ($\%$) on \textbf{Office31}. Best average HOS in \textbf{bold} and second best with an \underline{underline}.}
  \resizebox{0.99\textwidth}{!}{%
  \begin{threeparttable}
  \begin{tabular}{c|ccc|ccc|ccc|ccc|ccc|ccc|ccc}
    \Xhline{1px}
  {\multirow{2}{*}{Approach}} & \multicolumn{3}{c|}{Amazon$\to$DSLR} & \multicolumn{3}{c|}{Amazon$\to$Webcam} & \multicolumn{3}{c|}{DSLR$\to$Amazon} & \multicolumn{3}{c|}{DSLR$\to$Webcam} & \multicolumn{3}{c|}{Webcam$\to$Amazon} & \multicolumn{3}{c|}{Webcam$\to$DSLR}  &\multicolumn{3}{c}{Avg}\\
                              &OS*    & UNK    & \textbf{HOS}  & OS*    & UNK    & \textbf{HOS}       & OS*    & UNK    & \textbf{HOS}    & OS*    & UNK    & \textbf{HOS}    & OS*    & UNK    & \textbf{HOS}   & OS*    & UNK    & \textbf{HOS}    & OS*    & UNK    & \textbf{\textbf{HOS}}   \\
  \hline
  OSBP                  & 89.7 & 66.3 &76.2 & 89.1 &66.2 &75.9 & 62.1 &78.8 &69.4 & 79.8 &86.2 &82.9 &76.9 &66.9 &71.6 & 93.6 &77.1 &84.6 &81.9 & 73.6 &\cellcolor[HTML]{EFEFEF} 76.8 \\
  STA                  & 97.3 &75.4 & 85.0 & 99.0 & 71.7 & 83.2 & 91.5 & 70.7 & 79.7 & 99.7 &61.0 &75.7 & 93.4 & 65.4 & 77.0 & 100.0 & 55.4 &71.3 &96.8 &66.6 &\cellcolor[HTML]{EFEFEF} 78.7\\
  ROS                  &  66.5 & 86.3 & 75.1 &38.4 &71.7 &50.0 &66.3 &79.8 &72.4 &87.9 & 82.2 & 90.0 &66.5 & 87.0 &75.4 &95.1 &77.7 &85.5 &70.1 &80.8 &\cellcolor[HTML]{EFEFEF} 74.7\\
  PGL                  &  80.1 &65.1 & 71.8 & 90.1 & 68.3 & 77.7 & 62.0 & 58.0 & 60.0 & 87.1 & 65.3 & 74.6 & 69.6 & 59.2 & 64.0 & 77.3 & 61.8 & 68.7 &77.7 & 63.0 &\cellcolor[HTML]{EFEFEF} 69.5\\
  DCC                  &  98.7 & 89.8 & 94.0 & 93.9 & 92.8 & 93.3 & 94.5 & 75.2 &83.8 & 100.0 & 92.8 &96.3 &95.2 &81.2 &87.6 &100.0 & 89.2 &94.3 &97.1 & 86.8 & \cellcolor[HTML]{EFEFEF} 91.6\\
  DANCE                  &  84.1 & 33.1 & 47.5 & 91.5 & 55.4 & 69.0 & 67.5 & 63.6 & 65.5 & 97.9 & 55.0 & 70.4 & 92.0 & 47.4 & 62.5 &100.0 & 48.0 & 64.9 & 88.8 & 50.4 &\cellcolor[HTML]{EFEFEF} 63.3\\
  OVANet                  &92.3  &89.7 &91.0 & 91.6  &91.8 & 91.7 & 51.7  & 99.3  & 68.0 & 96.9  & 100.0 & 98.4 & 86.1 & 96.3 & 90.9 & 100.0 & 85.2 & 92.0 &86.4  & 93.7 &\cellcolor[HTML]{EFEFEF} 88.7\\
  GATE\tnote{$\dagger$ }                  & -  &-  &88.4  &-  &- & 86.5 & - &-  & 84.2  &- &- & 95.0   &-  &-  & 86.1 &- &- & 96.7 &-  &-  & \cellcolor[HTML]{EFEFEF} 89.5 \\
  UADAL                  & 81.2 & 88.6 & 84.7 & 84.3 & 96.7 & 90.1   & 72.6 & 90.0 & 80.4 & 100.0 & 92.5 & 96.1 & 72.8 & 92.4 & 81.4 & 100.0 & 98.0 & 99.0 &85.2 &93.0 & \cellcolor[HTML]{EFEFEF} 88.6\\
  ANNA                  &94.0 & 73.4 & 82.4  &94.6 & 71.5 & 81.5  &73.7 & 82.1 & 77.7  &99.5 & 97.0 & 98.2  &73.7 & 82.6 & 77.9 &100.0 & 89.9 & 94.7 & 89.3 & 82.8 &\cellcolor[HTML]{EFEFEF} 85.4\\
  GLC                  &  85.3 &90.6 &87.9 & 86.8 &93.1 &89.8 & 92.3 & 98.0 & 95.1 & 94.0 &96.4 & 95.2 & 91.9 &97.9 & 94.8 & 98.7 & 96.9 & 97.8 &91.5 &95.5 &\cellcolor[HTML]{EFEFEF} \underline{93.4}\\
  \hhline{|-|-|-|-|-|-|-|-|-|-|-|-|-|-|-|-|-|-|-|-|-|-|}
  DSD (Ours)                 &  96.9 & 91.3 & 94.0 & 91.2 & 94.4 & 92.8 &91.5 & 97.0 &94.1 &97.9 &91.9 &94.8 &93.4 & 95.3 & 94.4 & 99.1 & 91.3 & 95.0 & 95.0 & 93.5 & \cellcolor[HTML]{EFEFEF} \textbf{94.2}\\
  \Xhline{1px}
\end{tabular}

\begin{tablenotes}
  \footnotesize
  \item[$\dagger$ ] Cited from \cite{chen2022geometric}.
  \end{tablenotes}
\end{threeparttable}
  }
  
\label{tab:osda_office31}
\end{table*}

\begin{table*}[htpb]   \centering
\caption{Results ($\%$) on \textbf{OfficeHome}. `Threshold-Free' means no threshold is required. Best average HOS in \textbf{bold} and second best with an \underline{underline}.}
  \resizebox{0.99\textwidth}{!}{
  \begin{threeparttable}
  \begin{tabular}{cccc|ccc|ccc|ccc|ccc|ccc|ccc}
    \Xhline{1px}
  \multirow{2}{*}{Approach} & \multicolumn{3}{c|}{\multirow{2}{*}{Threshold-Free}} & \multicolumn{3}{c|}{Art$\to$Clipart} & \multicolumn{3}{c|}{Art$\to$Product} & \multicolumn{3}{c|}{Art$\to$Real-World} & \multicolumn{3}{c|}{Clipart$\to$Art} & \multicolumn{3}{c|}{Clipart$\to$Product} & \multicolumn{3}{c}{Clipart$\to$Real-World}\\
                          & \multicolumn{3}{c|}{}                           & OS*    & UNK    & \textbf{HOS}    & OS*    & UNK    & \textbf{HOS}    & OS*    & UNK    & \textbf{HOS}    & OS*    & UNK    & \textbf{HOS}    & OS*    & UNK    & \textbf{HOS}    & OS*    & UNK    & \textbf{HOS}\\
                          \hline
    OSBP                  & \multicolumn{3}{c|}{\xmark} &46.8 &77.2 & 58.3 &55.3 &72.0 & 62.6 &75.8 &55.6 &64.2  &52.6 &61.8 &56.8 &59.2 &47.1 &52.5 &67.7 &63.5 & 65.5\\
    STA                  & \multicolumn{3}{c|}{\cmark} &60.0 &56.2 &58.0 & 83.1 & 47.1 &60.1 & 90.6 &49.2 &63.8 & 68.9 &63.0 & 65.8 & 74.5 &49.4 & 59.4 & 77.7 & 51.4 &61.9\\
    ROS                  & \multicolumn{3}{c|}{\xmark} & 48.7 & 73.0 & 58.4  &64.1 & 66.0 & 65.0 & 73.8 & 62.6 & 67.7 & 52.6 &64.4 & 57.9 &59.8 &53.9 &56.7 & 58.2 & 76.0 & 66.0 \\
    PGL                  & \multicolumn{3}{c|}{\xmark} &63.8 & 51.3 & 56.9 & 80.2 & 58.4 & 67.6 & 88.7 & 63.4 & 73.9 & 71.0 & 59.0 & 64.4 & 74.1 & 56.4 & 64.1 & 81.4 & 59.5 & 68.8\\
    DCC                  & \multicolumn{3}{c|}{\cmark} &56.7 & 69.3 & 62.3 & 78.9 & 67.2 & 72.6 & 82.2 & 66.8 & 73.7 &54.1 & 56.1 & 55.1 &67.8 & 73.9 & 70.7 &  82.7 & 76.6 & 79.5\\
    DANCE                  & \multicolumn{3}{c|}{\xmark} & 44.7 & 65.1 & 53.0 & 60.3 & 60.0 & 60.1 & 80.7 & 68.9 & 74.3 & 45.6 & 74.7 & 56.6 & 64.5 & 68.7 & 66.5 & 39.7 & 86.7 & 54.4\\
    OVANet                  & \multicolumn{3}{c|}{\cmark} &42.6 &81.4  & 55.9  &72.3 & 65.6 & 68.8  &86.1   & 64.1  & 73.5  &48.7  &  83.4  & 61.5  &63.1  &73.9  &  68.1  &70.1   & 78.1  & 73.9\\
    GATE\tnote{$\dagger$ }                  & \multicolumn{3}{c|}{\cmark} & -    & -    & 63.8 & -    & -    & 70.5  & -   & -   & 75.8 &-     & -    & 66.4 & -    & -    & 67.9 & -    & -    & 71.7\\
    UADAL                  & \multicolumn{3}{c|}{\xmark} & 42.6 & 71.3 & 53.4 & 68.6 & 74.6 & 71.5 &85.7 & 73.2 & 78.9 & 52.1 & 82.5 &63.9 & 59.9 & 74.8 & 66.5 & 65.4 & 83.7 & 73.4\\
    ANNA                  & \multicolumn{3}{c|}{\xmark}  & 58.5 &72.3 &64.7 & 67.6 & 70.8 &69.2 & 72.7 &75.3 &74.0 & 48.9 &82.5 &61.4 &61.7 &69.9 &65.6 &68.4 & 76.4 & 72.2\\
    GLC                  & \multicolumn{3}{c|}{\xmark} & 62.6 & 69.3 & 65.8 & 75.6 & 73.6 & 74.6 & 81.4 & 80.2 & 80.8 & 71.4 & 34.4 & 46.4 & 77.9 & 76.2 & 77.0 & 82.1 & 82.2 & 82.1 \\
    \hhline{|-|-|-|-|-|-|-|-|-|-|-|-|-|-|-|-|-|-|-|-|-|-|}
    DSD (Ours)                & \multicolumn{3}{c|}{\cmark} & 49.5 & 78.1 & 60.6 & 63.6 & 78.7 & 70.3 & 82.3 & 75.6 & 78.8 & 65.4 & 69.4 & 67.3 & 62.5 & 82.6 & 71.2 & 71.8 & 79.9 & 75.6\\
    \hhline{|-|-|-|-|-|-|-|-|-|-|-|-|-|-|-|-|-|-|-|-|-|-|}
  {\multirow{2}{*}{Approach}} & \multicolumn{3}{c|}{Product$\to$Art} & \multicolumn{3}{c|}{Product$\to$Clipart} & \multicolumn{3}{c|}{Product$\to$Real-World} & \multicolumn{3}{c|}{Real-World$\to$Art} & \multicolumn{3}{c|}{Real-World$\to$Clipart} & \multicolumn{3}{c|}{Real-World$\to$Product}  &\multicolumn{3}{c}{Avg}\\
                             &OS*    & UNK    & \textbf{HOS}  & OS*    & UNK    & \textbf{HOS}       & OS*    & UNK    & \textbf{HOS}    & OS*    & UNK    & \textbf{HOS}    & OS*    & UNK    & \textbf{HOS}   & OS*    & UNK    &\textbf{HOS}    & OS*    & UNK    & \textbf{HOS}   \\
  \hline
  OSBP                  & 53.2& 64.5& 58.3& 49.6& 45.2& 47.3& 63.3 &70.1 &66.6 & 65.8 & 43.1 &52.1 &59.2& 34.1& 43.2 & 80.3 & 51.8 &  63.0 &60.7 &57.2 &\cellcolor[HTML]{EFEFEF} 57.5\\
  STA                   & 70.2 & 62.3 & 66.0 & 57.3 & 54.8 & 56.0 & 86.2 & 57.7 & 69.2 & 77.9 & 53.4 & 63.4 & 59.5 & 43.4 & 50.2 & 86.3 &41.1 &55.7 & 74.4 & 52.4 &\cellcolor[HTML]{EFEFEF} 60.8\\
  ROS                  & 48.5 &63.9 &55.1 & 44.3 &73.5 &55.3 & 66.3 & 83.3 & 73.8 &68.1 &59.9 &63.7 &48.5 & 71.9 &57.9 &71.3 &46.2 &56.1 &58.7 & 66.2 &\cellcolor[HTML]{EFEFEF} 61.1\\
  PGL                  & 72.0 & 60.6 & 65.8 & 63.2 & 52.3 & 57.2 & 82.5 & 60.3 & 69.7 & 73.7 & 59.2 &65.7 &63.0 &51.0 &56.4 &83.6 &60.1 &69.9  & 74.8 &57.6 &\cellcolor[HTML]{EFEFEF} 65.0\\
  DCC                  &55.1  &78.6 &64.8 &51.7 &73.1 &60.6 &76.6 &73.9 &75.2 & 75.5 &55.9 &64.2 &57.4 &65.0 &60.9 &79.2 &74.0 &76.5 & 68.2 & 69.2 &\cellcolor[HTML]{EFEFEF} 68.0\\
  DANCE                  & 59.7 & 54.0 & 56.7 & 51.3 & 63.4 & 56.7 & 74.8 & 66.1 & 70.1 & 72.4 & 30.0 & 42.5 & 64.8 & 30.9 & 41.8 & 81.1 & 43.7 & 56.8 & 61.6 & 59.4 &\cellcolor[HTML]{EFEFEF} 57.5\\
  OVANet                  &50.1  & 84.8  &  63.0 &37.3     & 83.5  & 51.6  &77.4     & 74.8  & 76.1  &72.3  & 69.0  & 70.6  &46.5 & 70.7 & 56.1 & 82.5 & 56.1 & 66.8 & 62.4 & 73.8 &\cellcolor[HTML]{EFEFEF} 65.5\\
  GATE\tnote{$\dagger$ }                  & -    & -    & 67.3 & -    & -    & 61.5  & -   & -   & 76.0 &-     & -    & 70.4 & -    & -    & 61.8 & -    & -    & 75.1 &-   &-   & \cellcolor[HTML]{EFEFEF} \underline{69.0} \\
  UADAL                  & 59.6 & 81.3 & 68.8 & 32.5 & 75.1 & 45.3 & 79.4 & 78.8 & 79.1 & 73.6 & 70.7 & 72.2 & 38.3 & 73.9 & 50.5 & 80.8 & 70.8 & 75.5 & 61.5 & 75.9 & \cellcolor[HTML]{EFEFEF} 66.6  \\
  ANNA                   & 52.0 & 80.9 & 63.3 & 51.1 &79.6 &62.2 & 67.8 &79.6 &73.2 &58.8 & 83.5 & 69.0 & 56.1 & 75.7 & 64.4 & 73.0 & 85.6 & 78.8 &61.4 &77.7 &\cellcolor[HTML]{EFEFEF} 68.2\\
  GLC                  & 61.5 & 84.5 & 71.2 & 48.5 & 91.6 & 63.4 & 75.1 & 78.2 & 76.6 & 75.9 & 35.4 & 48.2 & 36.1 & 90.2 & 51.6 & 83.4 & 81.2 & 82.3 &69.3 &73.1 & \cellcolor[HTML]{EFEFEF} 68.3 \\
  \hhline{|-|-|-|-|-|-|-|-|-|-|-|-|-|-|-|-|-|-|-|-|-|-|}
  DSD (Ours)                 &  61.5 & 82.1 & 70.3 & 42.5 & 83.8 & 56.4 &78.8 & 76.6 & 77.7 & 73.7 & 71.2 & 72.5 & 47.0 & 80.1 & 59.3 & 76.1 & 79.8 & 77.9 &64.6 &78.2 & \cellcolor[HTML]{EFEFEF} \textbf{69.8}\\
  \Xhline{1px}
\end{tabular}
\begin{tablenotes}
\footnotesize
\item[$\dagger$ ] Cited from \cite{chen2022geometric}.
\end{tablenotes}
\end{threeparttable}
}
\label{tab:osda_officehome}
\end{table*}

\subsection{Experiment Setup}
\begin{itemize}
 \item \textbf{Datasets.} Three typical datasets were used:
\begin{enumerate}
\item \textbf{Office31} \cite{saenko2010adapting}, which contains 4,652 images from 31 classes in three different domains: Amazon, Webcam, and DSLR.
\item \textbf{OfficeHome} \cite{venkateswara2017deep}, which contains 15,500 images from 65 classes in four different domains: Art, Clipart, Product, and Real-World.
\item \textbf{VisDA} \cite{visda2017}, which contains over 200,000 images from 12 classes in different domains.
\end{enumerate}
$|C_s|/|y_{unk}|$ class split was $10/11$ on Office31, $25/40$ on OfficeHome, and $6/6$ on VisDA, following previous studies \cite{jang2022unknown, sanqing2023GLC}.

\item \textbf{Evaluation metric.} As in earlier studies, HOS \cite{bucci2020effectiveness}, which calculates the harmonic mean of the average accuracies of known classes (OS*) and the accuracy of unknown class (UNK):
\begin{align}
  \text{HOS} = \frac{2\text{OS*}\times \text{UNK}}{\text{OS*}+\text{UNK}},
\end{align}
was used as the evaluation metric.

\item \textbf{Baselines.} Our proposed DSD was compared with existing OSDA approaches, e.g., OSBP \cite{saito2018open}, Separate to Adapt (STA) \cite{liu2019separate}, ROS \cite{bucci2020effectiveness}, PGL \cite{luo2020progressive}, UADAL \cite{jang2022unknown}, ANNA \cite{li2023adjustment}, and multiple UniDA approaches in the OSDA setting, e.g., DCC \cite{li2021domain}, Domain Adaptative Neighborhood Clustering via Entropy Optimization (DANCE) \cite{saito2020universal}, One-Versus-All Network (OVANet) \cite{saito2021ovanet}, Geometric Anchor-guided Adversarial and Contrastive Learning Framework with Uncertainty Modeling (GATE) \cite{chen2022geometric}, and GLC \cite{sanqing2023GLC}.

\item \textbf{Implementation details.} For a fair comparison, we implemented all approaches using the ImageNet \cite{deng2009imagenet} pre-trained Data-efficient image Transformers Small (Deit-S) \cite{touvron2021training} backbone network, except GATE (since we were not able to access the open-source code of GATE \cite{chen2022geometric}, we used the results published in the original paper, which were based on ResNet50 \cite{he2016deep}, with similar number of parameters and run-time memory cost to Deit-S) and ANNA (ANNA \cite{li2023adjustment} extracts pixel-level features using ResNet50 \cite{he2016deep} and we implemented it following the original settings). Our proposed DSD used Deit-S as the backbone network, keeping the same MoE structure configuration as previous work \cite{li2023sparse}. Model optimization was done using Nesterov momentum SGD with momentum 0.9 and weight decay 1e-3. Batch size 64, learning rate 1e-4, $m=0.99$, $N=6$, $K=2$, and $\gamma=100$ were used for all datasets.

\end{itemize}

\subsection{Experiment Results}

\begin{table}[htpb]
  \centering
  \caption{Results ($\%$) on \textbf{VisDA}. Best HOS in \textbf{bold} and second best with an \underline{underline}.}
  \resizebox{\columnwidth}{!}{%
  \begin{threeparttable}
  \begin{tabular}{c|ccc|c|ccc}
    \Xhline{1px}
  {\multirow{2}{*}{Approach}} & \multicolumn{3}{c|}{Sy$\to$Re} &\multirow{2}{*}{Approach} & \multicolumn{3}{c}{Sy$\to$Re} \\
                             &OS*    & UNK    & \textbf{HOS}              &                    & OS*    & UNK    & \textbf{HOS}          \\
  \hline
  OSBP                  & 25.2  & 39.1 & \cellcolor[HTML]{EFEFEF} 30.7  &OVANet                  &41.8  & 78.7  & \cellcolor[HTML]{EFEFEF} 54.6\\
  STA                  & 65.6 & 80.9  &\cellcolor[HTML]{EFEFEF} \underline{72.4} & GATE\tnote{$\dagger$ }                  &  -  &-  & \cellcolor[HTML]{EFEFEF} 70.8\\
  ROS                  & 36.5 &87.0 &\cellcolor[HTML]{EFEFEF} 51.4 & UADAL                  &  50.9 & 93.3 & \cellcolor[HTML]{EFEFEF} 65.9\\
  PGL                  &  74.4 & 34.1 & \cellcolor[HTML]{EFEFEF} 46.8 & ANNA                  &  37.9 & 83.5 & \cellcolor[HTML]{EFEFEF} 52.1\\
  DCC                  &  56.3 & 86.1 & \cellcolor[HTML]{EFEFEF} 68.1 & GLC                  &  55.9 & 93.0 & \cellcolor[HTML]{EFEFEF} 69.8\\
  \hhline{|~|~|~|~|-|-|-|-|}
  DANCE                  &  58.7 & 92.2 & \cellcolor[HTML]{EFEFEF} 71.7 & DSD (Ours)                 &  73.5 &77.7 &\cellcolor[HTML]{EFEFEF} \textbf{75.5}\\
  \Xhline{1px}
\end{tabular}%
\begin{tablenotes}
  \footnotesize
  \item[$\dagger$ ] Cited from \cite{chen2022geometric}.
  \end{tablenotes}
\end{threeparttable}
  }
  
\label{tab:osda_visda}
\end{table}

Tables~\ref{tab:osda_office31}-\ref{tab:osda_visda} show the performance of all algorithms on the three datasets, respectively. On average, our proposed DSD achieved the best performance on all datasets, without using any threshold.

\begin{enumerate}
  \item \textbf{Office31} (Table \ref{tab:osda_office31}). Our approach achieved a better balance between OS* and UNK. DSD demonstrated good recognition performance for both known and unknown classes, with an average HOS improvement of $0.8\%$ over GLC and $2.6\%$ over DCC.
  \item \textbf{OfficeHome} (Table \ref{tab:osda_officehome}). Our approach improved UNK significantly, outperforming GLC by $5.1\%$. It also achieved the best average HOS, outperforming GATE, GLC and UADAL by $0.8\%$, $1.5\%$ and $3.2\%$, respectively.
  \item \textbf{VisDA} (Table \ref{tab:osda_visda}). Our approach demonstrated superior performance in differentiating known samples from unknown ones, yielding $3.1\%$ and $5.7\%$ HOS gains over STA and GLC. In contrast, other approaches biased towards higher UNK and lower OS*, suggesting they are prone to misclassify known class samples as unknown.
\end{enumerate}

Generally, DSD achieved a more balanced UNK and OS*, indicating that it can well classify known classes in the target domain, while simultaneously identifying the unknown ones. Thanks to its threshold-free nature, DSD achieved \textit{best performance across all three datasets}, compared to previous approaches that require carefully tuned hyper-parameters per dataset, demonstrating good generalization and requiring fewer empirical choices of hyper-parameters.

\begin{figure}[htpb]  \centering
  \includegraphics[width=1.0\columnwidth]{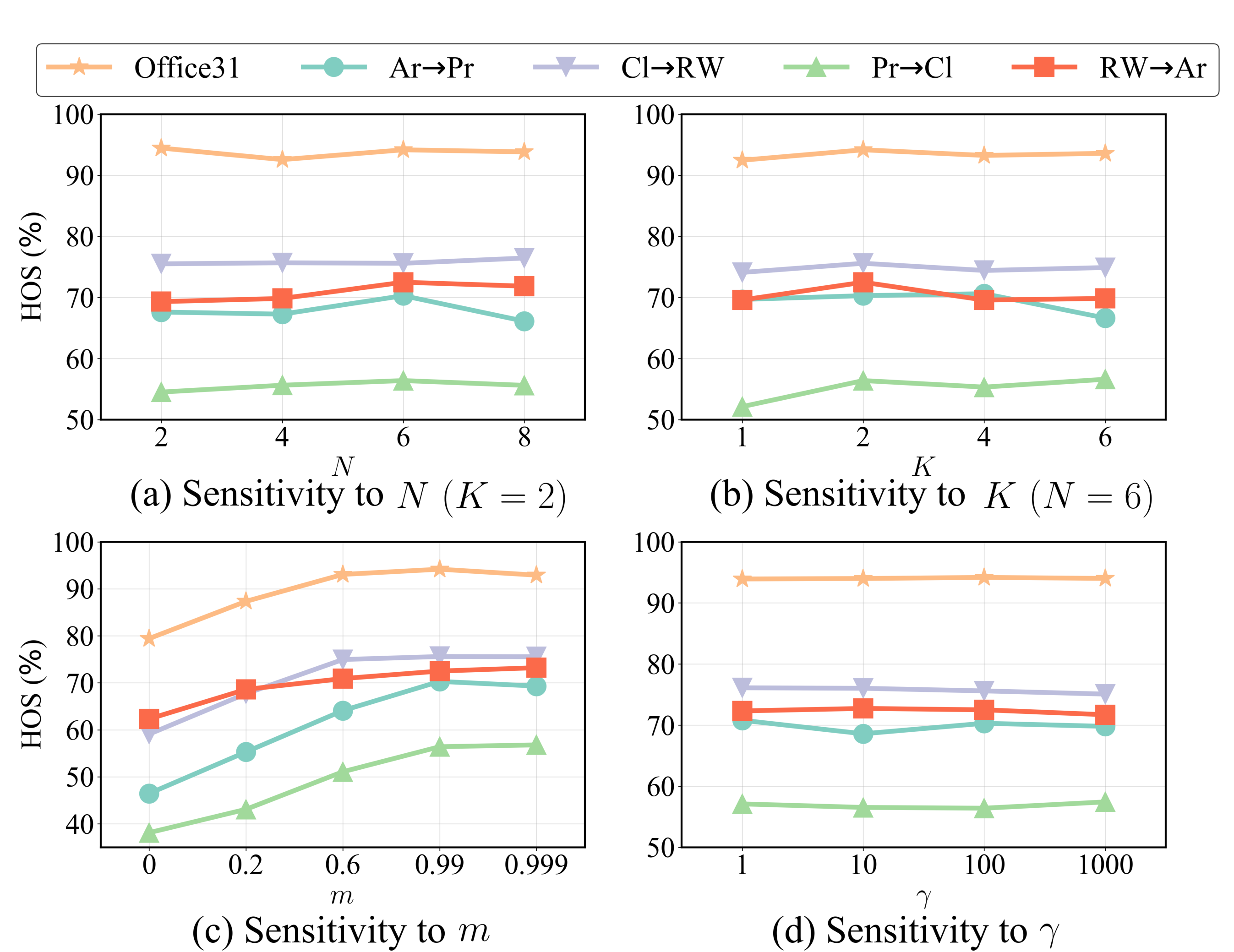}
  \caption{Hyper-parameters analysis on Office31 and Art$\to$Product, Clipart$\to$Real-World, Product$\to$Clipart and Real-World$\to$Art on OfficeHome. (a) $N$, the total number of experts; (b) $K$, the number of experts selected during each routing step; (c) $m$ in Eq.~(\ref{eq:m_ins}); and, (d)  $\gamma$.}
\label{fig:sens}
\end{figure}

\subsection{Ablation study of the Graph Router}

The router is very important to the performance of MoE. Many different approaches have been proposed. To validate the efficacy of our proposed Graph Router, we used the widely-adopted Cosine Router \cite{li2023sparse} to replace Graph Router in the GRMoE layer.
The Cosine Router (CR) is
\begin{align}
   \text{CR}(\bm{x}) = \frac{\text{E}^T\text{W}\bm{x}}{\tau \|\text{W}\bm{x}\| \|\text{E} \|},
\end{align}
where $\text{W}$ is a projection matrix, $\text{E}$ a learned embedding modulated by $\tau$, and $\|\cdot\|$ indicates the modulus of the vector.

 We have also studied to replace the GAT in Graph Router with MHSA and graph convolution layer (GCN). The results are shown in Table~\ref{tab:router}. Clearly, the Graph Router gets the best performance across all three datasets. Unlike Cosine Router that processes tokens independently, the Graph Router aggregates nearby information from adjacent tokens, enhancing the ability of experts to catch spatial critical features. As MHSA integrates information from different tokens, it treats all tokens equally, thus the class token representing the routing feature cannot capture the global critical information, leading to a dramatic performance decline. Meanwhile, the GCN has a poorer ability to extract information comparing to the GAT, with a slight HOS decrease.

\begin{table}[h]  \centering
  \caption{Comparison of different routers' HOS~(\%) on the three datasets.}
  \addtolength{\tabcolsep}{3.0pt}
  \resizebox{0.47\textwidth}{!}{
      \begin{tabular}{c|ccc}
          \Xhline{1px}
          Router & Office31 & OfficeHome & VisDA \\
          \hline
          Cosine Router  & 93.2   & 67.3  & 72.4 \\
          Graph Router (MHSA)  & 76.9   & 60.7  & 67.5 \\
          Graph Router (GCN)  & 92.1   & 67.6 & 74.4 \\
          \textbf{Graph Router (Ours)}  & \textbf{94.2}  & \textbf{69.8}  & \textbf{75.5} \\
          \Xhline{1px}
      \end{tabular}  }   
      
      \label{tab:router}
\end{table}

\subsection{Hyper-parameters Analysis}

To validate the sensitivity to different hyper-parameters of DSD, we conducts following experiments:

\begin{enumerate}
    \item Effect of $N$ and $K$ in MoE. The GRMoE layer includes two important hyper-parameters: $N$, the total number of experts, and $K$, the number of experts selected during each routing step. We studied the effect of each, by fixing the other. Fig.~\ref{fig:sens}(a) and (b) show the results. Despite the variations in $N$ and $K$, the results exhibited remarkable stability, indicating the robustness of our approach against its hyper-parameters. It further demonstrates that the remarkable performance of DSD can be traced back to the strategic exploitation of inconsistencies between two spaces, rather than an increase of model capacity.

    \item Effect of $m$ in momentum update. DSD uses two separate memories to store the routing features and the image features. A momentum-based update strategy is employed during training to update these features. The influence of hyper-parameter $m$ on the performance is illustrated in Fig.~\ref{fig:sens}(c). A larger $m$ generally resulted in better performance, as it stabilizes the training process, which is particularly beneficial in handling the randomness in clustering unknown class samples.

    \item Effect of $\gamma$ in the loss function. $\gamma$ controls the contribution of $\mathcal{L}_{\text{blc}}$ to the loss function, which promotes balanced use of the experts. Fig.~\ref{fig:sens}(d) shows the effect of $\gamma$. Generally speaking, the HOS is stable across different values of $\gamma$.
\end{enumerate}

\begin{figure}[ht]  \centering
  \includegraphics[width=\linewidth]{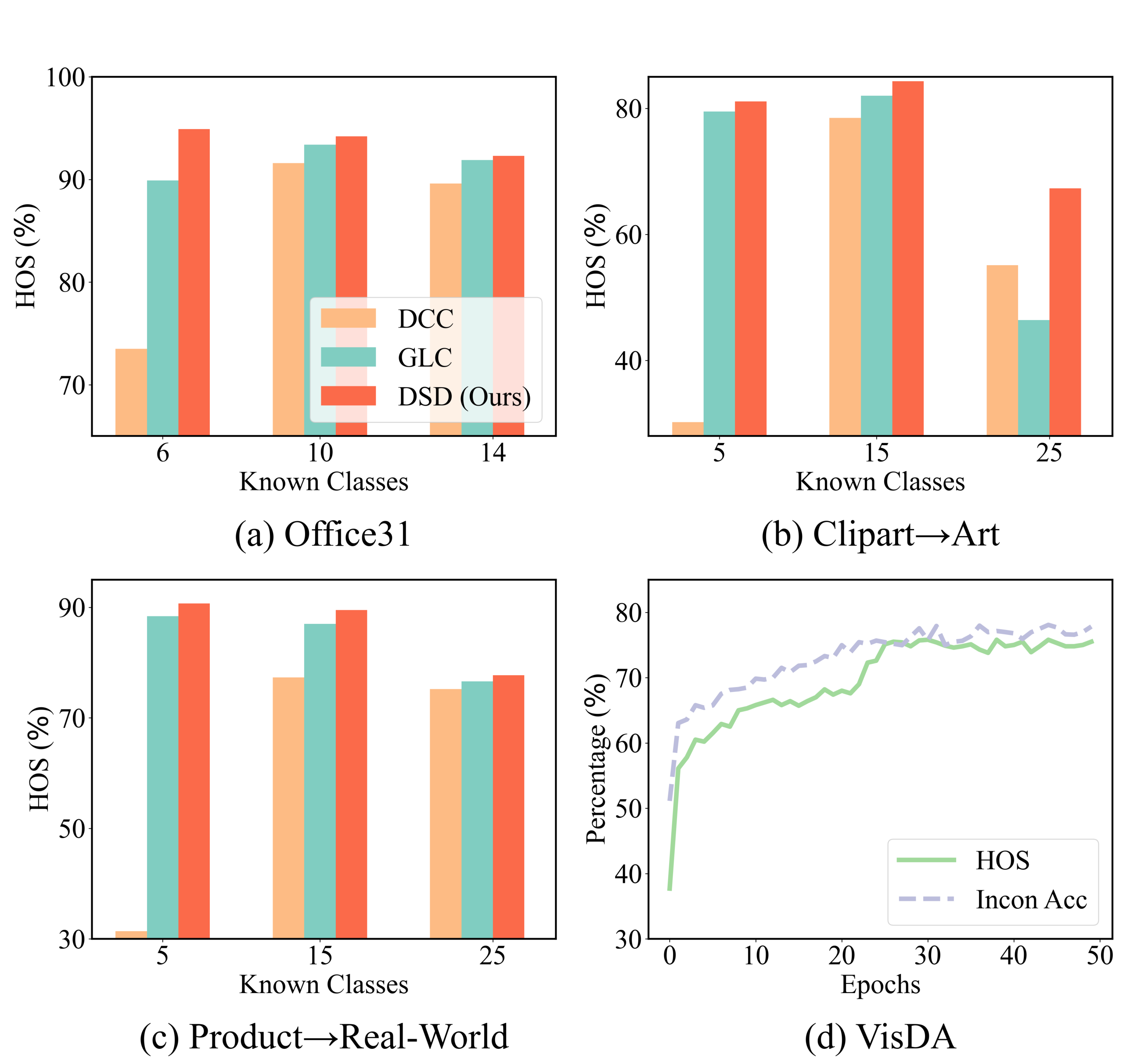}
  \caption{(a)-(c) Performance of DCC, GLC and DSD (Ours) on Office31 and Clipart$\to$Art, Product$\to$Real-World on OfficeHome with different numbers of known classes. (d) The learning curves on VisDA, where `Incon Acc' denotes the percentage of samples with inconsistent pseudo-labels being from unknown classes.} \label{fig:quan}
\end{figure}

\begin{figure*}[h]  \centering
  \includegraphics[width=\linewidth]{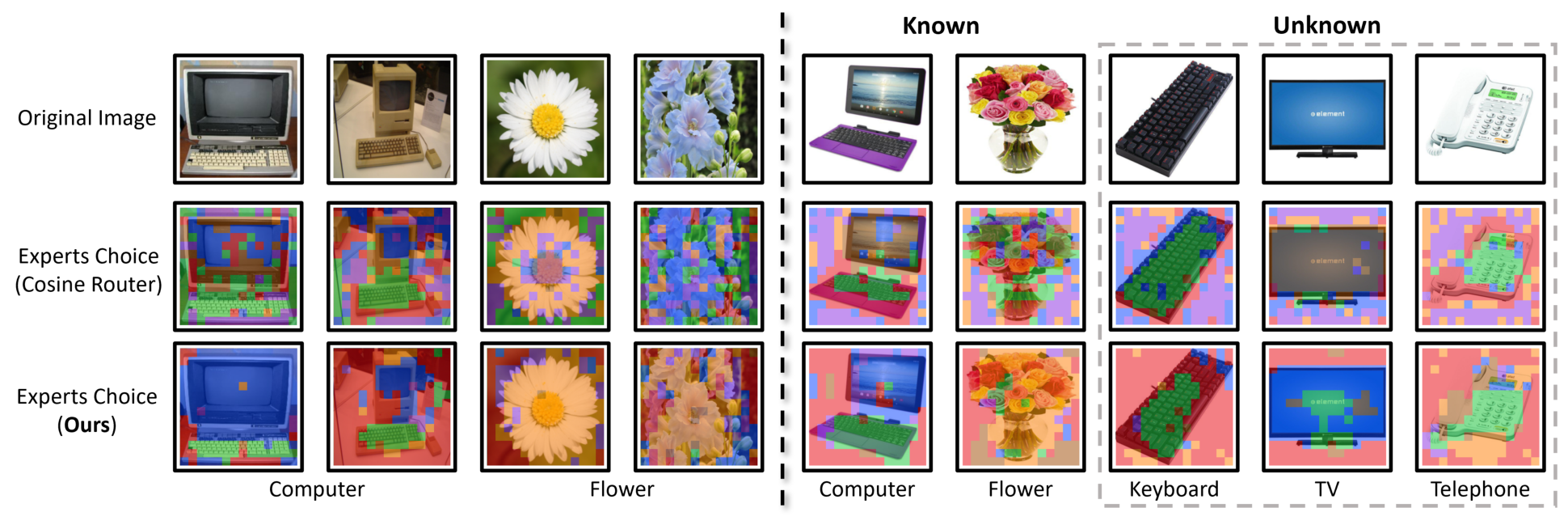}
  \caption{Visualization of the experts choice by Cosine Router and Graph Router on OfficeHome. Left: samples from the source domain (Real-World); Right: samples from the target domain (Product). Different colors represent different dominate experts of different patches. Gray dashed borders indicate unknown class samples.} \label{fig:choice}
\end{figure*}

\subsection{Quantization Analysis}

To demonstrate the robustness of our DSD, we varied the number of known classes, and Fig.~\ref{fig:quan}(a)-(c) show the HOS of different approaches on Office31 and Clipart$\to$Art and Product$\to$Real-World on OfficeHome, as the number of known classes varies. Our approach always achieved the best performance, showing its ability to distinguish known and unknown samples accurately.

We also show the learning curves of HOS and the accuracy of labeling the samples with inconsistent pseudo-labels as unknown class samples on VisDA in Fig.~\ref{fig:quan}(d). The recognition accuracy of unknown class samples gradually improved during the training process, leading to an increased HOS. This further demonstrates the effectiveness of DSD using the inconsistencies between two spaces to identify unknown class samples.

\subsection{MoE Layer Setting} \label{sec:moe_stru}

We used the recommended structure in \cite{li2023sparse}: the $9$-th and $11$-th layers of the original 12 layers Deit-S were replaced by GRMoE layers, and the routing feature space was obtained from the $11$-th layer's router output. Table~\ref{tab:moe} shows results of different MoE settings, including Last Layer (only replacing the $12$-th layer of Deit-S by GRMoE) and $9$-th Routing (the output of the $9$-th GRMoE layer as the routing feature space).

As Table~\ref{tab:moe} demonstrates, our setting yielded the best performance. Specifically, utilizing the output of the $9$-th layer as the routing space led to a significant decrease in performance, indicating that features generated in earlier layers are too general to identify unknown class samples.

The poorer performance observed in the Last Layer setting may be attributed to the fact that it is too close to the model's final image features, leading to reduced discriminability for unknown class samples.

\begin{table}[h]    \centering
   \caption{HOS (\%) on Office31, OfficeHome, and VisDA with different MoE settings.}
   \addtolength{\tabcolsep}{3.0pt}
   \resizebox{0.47\textwidth}{!}{
       \begin{tabular}{c|ccc}
           \Xhline{1px}
           Setting & Office31 & OfficeHome & VisDA \\
           \hline
           Last Layer  & 92.7   & 68.6  & 69.3 \\
           $9$-th Routing  & 84.3   & 61.8  & 62.6 \\
           \textbf{Our Setting}  & \textbf{94.2}  & \textbf{69.8}  & \textbf{75.5} \\
           \Xhline{1px}
       \end{tabular}    }    
       \label{tab:moe}
\end{table}

\subsection{Experts choice visualization}

We visualize the routing patterns of different class samples in the MoE layer in Fig.~\ref{fig:choice}. Samples with similar features in the source and target domains are routed through similar expert pathways. For instance, the blue-colored expert is responsible for distinguishing screens, and the green-colored expert for keyboards. This pattern exists not only in the known class `Computer' in the source domain but also in the unknown class samples in the target domain, validating the assumption that similar classes should also be close in the routing feature space. A close comparison of the Cosine Router and our Graph Router reveals a more consistent expert routing pattern for the latter. The proposed Graph Router exploits
the spatial relationships between nearby patches, unlike the per-patch routing approaches, leading to more consistent routing patterns and hence better performance.

\begin{figure}[htpb]  \centering
  \includegraphics[width=\columnwidth]{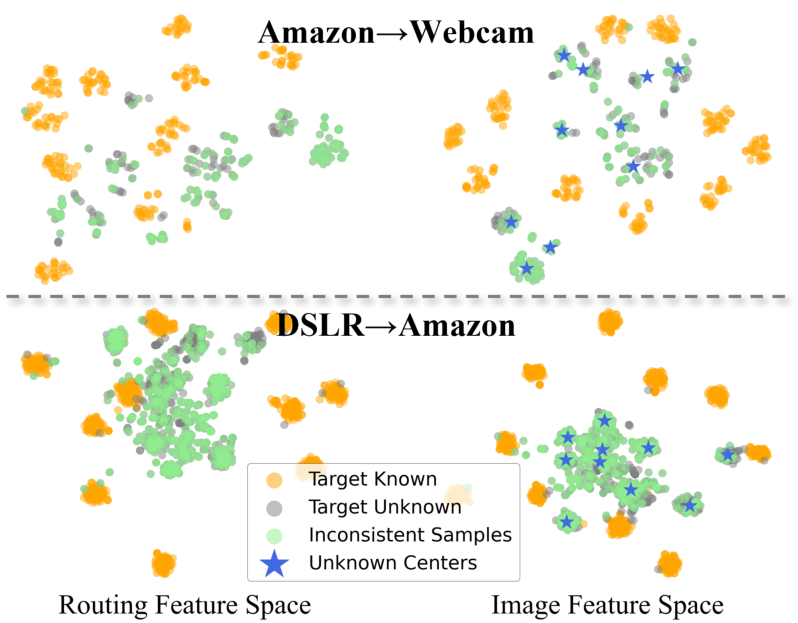}
  \caption{T-SNE visualization of the routing feature space (left) and the image feature space (right) in Amazon$\rightarrow$Webcam (top) and DSLR$\rightarrow$Amazon (bottom) on Office31.} 
  \label{fig:tsne} 
\end{figure}

\subsection{Feature distribution visualization}

To study the feature distribution, we used t-SNE \cite{van2008visualizing} to visualize both space features in Fig.~\ref{fig:tsne}.able concentration of samples inconsistent pseudo-labelsamong the unknown samples indicates the accuracy of DSD in identifying unknown samples, and hence the validity of inconsistencies as an unknown class measure. Furthermore, the clustered centroids resemble the distribution of the unknown class samples, implying their effectiveness in capturing the potential distribution of these samples.

\section{Conclusion and Future Work}
\label{sec:con}

This work has proposed a novel threshold-free DSD approach for OSDA, by considering the inconsistencies between the image feature space and the routing feature space in an MoE, enhanced by a Graph Router. DSD demonstrated promising performance in accurately identifying unknown class samples while maintaining reliable performance on known class samples, without using any thresholds. Comparisons with existing approaches demonstrated the robustness and versatility of our proposed approach, specifically shedding light on the significance of the Graph Router in leveraging spatial information among image patches. 

Our future research will explore the role of the routing feature space in other applications of MoE, such as multi task learning \cite{ma2018mmoe} and long tail problem \cite{Wang2021long}.

\bibliographystyle{IEEEtran}\bibliography{bibfile}

\end{document}